\pdfoutput=1
\documentclass[11pt]{article}

\usepackage[]{EMNLP2022}

\usepackage{times}
\usepackage{latexsym}

\usepackage[T1]{fontenc}
\usepackage[utf8]{inputenc}

\usepackage{microtype}

\usepackage{inconsolata}

\usepackage{multirow}
\usepackage{graphicx} 
\usepackage{amsmath}
\usepackage{amssymb} 
\usepackage{diagbox}
\usepackage{url}

\newcommand{\mname}{AccAlign }
\newcommand{\mnamec}{AccAlign}
\newcommand{\pname}{LaBSE }
\newcommand{\pnamec}{LaBSE}
\newcommand\numberthis{\addtocounter{equation}{1}\tag{\theequation}}

\title{Multilingual Sentence Transformer as A Multilingual Word Aligner}

\author{  Weikang Wang$^{1}$\footnotemark[1]\quad Guanhua Chen$^{2}$\thanks{~~The first two authors contribute equally.}~~\quad
  Hanqing Wang$^{1}$\quad
  Yue Han$^{1}$\quad
  Yun Chen$^{1}$\thanks{~~Corresponding author.}~~ \\
  $^1$Shanghai University of Finance and Economics \\
  $^2$Southern University of Science and Technology \\
  wwk@163.sufe.edu.cn \quad ghchen08@gmail.com \\
  \{whq,hanyue\}@163.sufe.edu.cn \quad yunchen@sufe.edu.cn \\ 
  }

\begin{document}
\maketitle
\begin{abstract}
Multilingual pretrained language models (mPLMs) have shown their effectiveness in multilingual word alignment induction. However, these methods usually start from mBERT or XLM-R. In this paper, we investigate whether multilingual sentence Transformer \pname is a strong multilingual word aligner. This idea is non-trivial as \pname is trained to learn language-agnostic sentence-level embeddings, while the alignment extraction task requires the more fine-grained word-level embeddings to be language-agnostic. We demonstrate that the vanilla \pname outperforms other mPLMs currently used in the alignment task, and then propose to finetune \pname on parallel corpus for further improvement. Experiment results on seven language pairs show that our best aligner outperforms previous state-of-the-art models of all varieties. In addition, our aligner supports different language pairs in a single model, and even achieves new state-of-the-art on zero-shot language pairs that does not appear in the finetuning process.
\end{abstract}

\section{Introduction}
Word alignment aims to find the correspondence between words in parallel texts \cite{brown1993mathematics}. It is useful in a variety of natural language processing (NLP) applications such as noisy parallel corpus filtering~\cite{kurfali-ostling-2019-noisy}, bilingual lexicon induction \cite{shi-etal-2021-bilingual}, code-switching corpus building~\cite{lee2019linguistically,lin2020pre} and incorporating lexical constraints into neural machine translation (NMT) models~\cite{hasler2018neural,chen2021lexically}.

Recently, neural word alignment approaches have developed rapidly and outperformed statistical word aligners like GIZA++~\cite{och-ney-2003-systematic} and fast-align \cite{dyer-etal-2013-simple}. Some works \cite{garg-etal-2019-jointly, li-etal-2019-word,zenkel2019adding,zenkel-etal-2020-end,  chen-etal-2020-accurate, zhang-van-genabith-2021-bidirectional,chen-etal-2021-mask} induce alignments from NMT model or its variants. 
However, these bilingual models only support the language pair involved in the training process. They also treat the source and target side differently, thus two models are required for bidirectional alignment extraction.  
Another line of works~\cite{jalili-sabet-etal-2020-simalign,dou-neubig-2021-word} build multilingual word aligners with contextualized embeddings from the multilingual pretrained language model \cite[mPLM]{mbert,conneau-etal-2020-unsupervised}. Thanks to the language-agnostic representations learned with multilingual masked language modeling task, these methods are capable of inducing word alignments even for language pairs without any parallel corpus. 

\begin{figure}[!t]
\centering
\includegraphics[width=1.0\columnwidth]{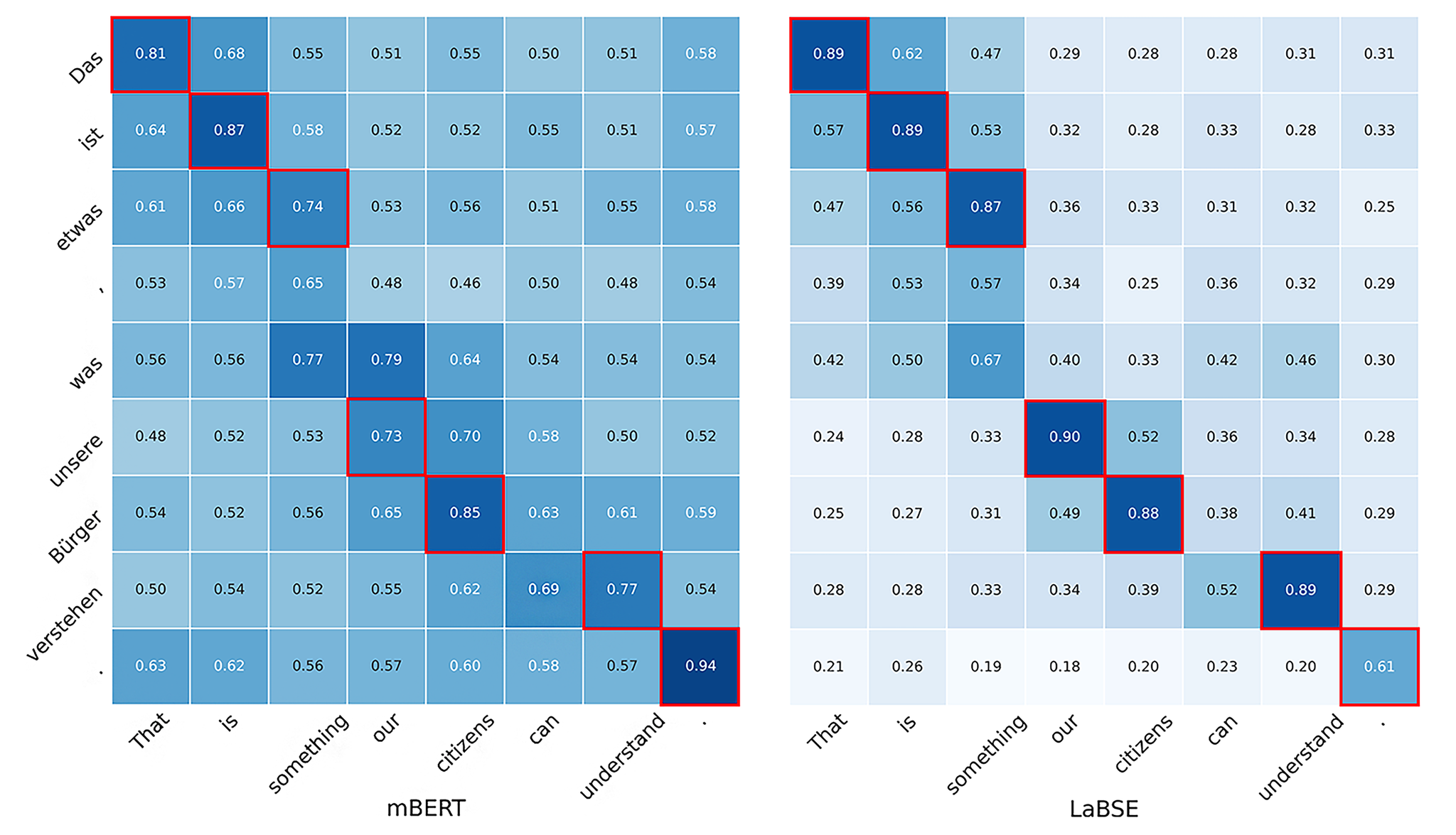}
\caption{Cosine similarities between subword representations in a parallel sentence pair from 8th layer of mBERT (left) and 6th layer of \pname (right). Red boxes denote the gold alignments.}
\label{fig:example}
\end{figure} 

Different from previous methods, in this paper we present \mnamec, a more accurate multilingual word aligner with the multilingual sentence Transformer LaBSE \cite[see Figure~\ref{fig:example}]{feng-etal-2022-language}. The LaBSE is trained on large scale parallel corpus of various language pairs to learn \emph{language-agnostic sentence embeddings} with contrastive learning. However, it is unclear whether \pname has learned \emph{language-agnostic word-level embeddings}, which is the key for the success of word alignment extraction. Specifically, we first direct induce word alignments from \pname and demonstrate that LaBSE outperforms other mPLMs currently used in the alignment task. This indicates that \pname has \emph{implicitly} learned language-agnostic word-level embeddings at some intermediate layer. Then we propose a simple and effective finetuning method to further improve performance. Empirical results on seven language pairs show that our best aligner outperforms previous SOTA models of all varieties. In addition, our aligner supports different language pairs in a single model, and even achieves new SOTA on zero-shot language pairs that does not appear in finetuning process.\footnote{Code is available at \url{https://github.com/sufenlp/AccAlign}.}

\section{\mname}
\subsection{Background: LaBSE}

LaBSE \cite{feng-etal-2022-language} is the state-of-the-art model for the cross-lingual sentence retrieval task. Given an input sentence, the model can retrieve the most similar sentence from candidates in a different language. \pname is first pretrained on a combination of masked language
modeling \cite{devlin-etal-2019-bert} and translation language modeling \cite{conneau2019cross} tasks. After that, it is effectively finetuned with contrastive loss on 6B parallel sentences across 109 languages. We leave the training detail of \pname in the appendix. However, as LaBSE does not include any word-level training loss when finetuning with contrastive loss, it is unclear whether the model has learned high-quality language-agnostic word-level embeddings, which is the key for a multilingual word aligner.

\subsection{Alignment Induction from LaBSE}
To investigate whether \pname is a strong multilingual word aligner, we first induce word alignments from vanilla \pname without any modification or finetuning. This is done by utilizing the contextual embeddings from \pnamec. Specifically, consider a bilingual sentence pair $\mathbf{x}=\langle x_1,x_2,...,x_n \rangle$ and $\mathbf{y}=\langle y_1,x_2,...,y_m \rangle$, we denote the contextual embeddings from \pname as $h_{\mathbf{x}}= \langle h_{x_1},...,h_{x_n} \rangle$ and $h_{\mathbf{y}}= \langle h_{y_1},...,h_{y_m} \rangle$, respectively. Following previous work~\cite{dou-neubig-2021-word,jalili-sabet-etal-2020-simalign}, we get the similarity matrix from the contextual embeddings:
\begin{equation}
    S=h_{\mathbf{x}}h_{\mathbf{y}}^{T}.
\end{equation}
The similarity matrix is normalized for each row to get $S_{\mathbf{x}\mathbf{y}}$. $S_{\mathbf{x}\mathbf{y}}$ is treated as the probability matrix as its i-th row represents the probabilities of aligning $x_i$ to all tokens in $\mathbf{y}$. The reverse probability matrix $S_{\mathbf{y}\mathbf{x}}$ is computed similarly by normalizing each column of $S$. Taking intersection of the two probability matrices yields the final alignment matrix:
\begin{equation}   
     A = (S_{\mathbf{x}\mathbf{y}}>c)*(S_{\mathbf{y}\mathbf{x}}^{T}>c),
     \label{eq:align}
\end{equation}
where $c$ is a threshold and $A_{ij}= 1$ indicates that $x_{i}$ and $y_{j}$ are aligned. The above method induces alignments on the subword level, which are converted into word-level alignments by aligning two words if any of their subwords are aligned following~\cite{zenkel-etal-2020-end,jalili-sabet-etal-2020-simalign}.

\subsection{Finetuning \pname for Better Alignments} 
\begin{table*}
\centering
  \resizebox{1.0\textwidth}{!}{
\begin{tabular}{llrrrrrrrrr}
\hline

Model         & Setting  &  & de-en & sv-en & fr-en & ro-en & ja-en & zh-en  & fa-en &avg \\ \hline
\multicolumn{10}{c}{Bilingual Statistical Methods}                                                                  \\ \hline  
fast-align~\cite{dyer-etal-2013-simple}    & \multirow{3}{*}{scratch}                  &  & 27.0   & -   & 10.5       & 32.1      &    51.1   &    38.1      & -  & -     \\
eflomal~\cite{ostling2016efficient}       &                   &  & 22.6  & -     & 8.2      & 25.1      & 47.5      & 28.7           & -  & -     \\
GIZA++~\cite{och-ney-2003-systematic}        &               &  & 20.6  & -    & 5.9      & 26.4      & 48.0      & 35.1            & - & -      \\ \hline
\multicolumn{10}{c}{Bilingual Neural Methods}                                                                     \\ \hline

MTL-FULLC-GZ~\cite{garg-etal-2019-jointly}
& \multirow{5}{*}{scratch}                            &  & 16.0  & -    & 4.6      & 23.1      & -      & -            & -   & -    \\
BAO-GUIDE~\cite{zenkel-etal-2020-end}     &                         &  & 16.3   & -     & 5.0      & 23.4      & -      & -          & -    & -   \\
SHIFT-AET~\cite{chen-etal-2020-accurate}     &                        &  & 15.4  & -    & 4.7      & 21.2      & -      & 17.2            & -  & -     \\
MASK-ALIGN~\cite{chen-etal-2021-mask}    &                          &  & 14.4  & -     & 4.4      & 19.5      & -      & 13.8           & -  & -     \\
BTBA-FCBO-SST~\cite{zhang-van-genabith-2021-bidirectional} &                         &  & 14.3  & -    & 6.7      & \textbf{18.5}      & -      & -            & -   & -    \\ \hline
\multicolumn{10}{c}{Multilingual Neural Methods}                                                \\ \hline
SimAlign~\cite{jalili-sabet-etal-2020-simalign}      & no ft                              &  & 18.8  & 11.2    & 7.6      & 27.2      & 46.6      & 21.6            & 32.7    & 23.7   \\ \hline
\multirow{3}{*}{AwesomeAlign~\cite{dou-neubig-2021-word}} & no ft                         &  & 17.4  & 9.7    &    5.6   & 27.9      & 45.6      & 18.1            & 33.0   & 22.5    \\
              &  self-sup ft                & & 15.9 & 7.9 & 4.4      & 26.2      & 42.4      & 14.9            & 27.1    & 19.8      \\
              & sup ft                   & & 15.2 & 7.2 & 4.0      & 25.5      & 40.6      & 13.4           & 25.8     & 18.8        \\ \hline
\multirow{3}{*}{\mname} &   no ft                       &  & 16.0  & 7.3    & 4.5      & 20.8      & 43.3      & 16.2            & 23.4    & 18.8   \\
             & self-sup ft                    &  & 14.3  & 5.8    & 3.9      & 21.6      & 39.2      & 13.0            & 22.6   & 17.2    \\
             & sup ft        &  & \textbf{\underline{13.6}}  & \textbf{\underline{5.2}}    & \textbf{\underline{2.8}}      & 20.8      & \textbf{\underline{36.9}}      & \textbf{\underline{11.5}}            & \textbf{\underline{22.2}}   & \textbf{\underline{16.1}}    \\ \hline
\end{tabular}
}
      \caption{AER comparison between \mname and the baselines on test set of $7$ language pairs. self-sup and sup mean finetuning the model with parallel corpus of self-supervised and human-annotated alignment labels, respectively. All multilingual methods are tested on zero-shot language pairs. }
      \label{tab:main}
\end{table*}

Inspired by \cite{dou-neubig-2021-word}, we propose a finetuning method to further improve performance given parallel corpus with alignment labels.

\begin{figure}[t]
\centering
\includegraphics[width=0.3\textwidth]{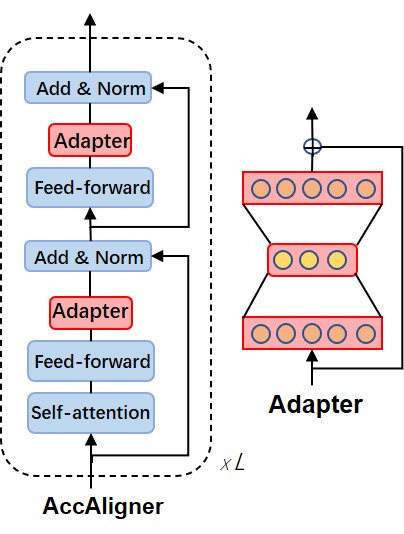}
\caption{The framework of adapter-based finetuning. The blue blocks are kept frozen, while the red adapter blocks are updated during finetuning. }
\label{fig:adapter}
\end{figure} 

\paragraph{Adapter-based Finetuning} Adapter-based finetuning \cite{houlsby2019parameter, bapna-firat-2019-simple, he-etal-2021-effectiveness} is not only parameter-efficient, but also benefits model performance, especially for low-resource and cross-lingual tasks~\cite{he-etal-2021-effectiveness}. 
Figure~\ref{fig:adapter} illustrates our overall framework, where the adapters are adopted from \cite{houlsby2019parameter}. For each layer of \pnamec, we introduce an adapter for each sublayer, which maps the input vector of dimension $d$ to dimension $m$ where $m< d$, and then re-maps it back to dimension $d$. Let $h$ and $h'$ denote the input and output vector, respectively. The output vector $h'$ is calculated as:
\begin{equation}
h^{'} = W_{up}\cdot \text{tanh}(W_{down}\cdot  h)  + h.
\end{equation}
Note that a skip-connection is employed to approximate an identity function if parameters of the projection matrices are near zero. During finetuning, only parameters of the adapters are updated. 

\paragraph{Training Objective} Let $\hat{A}$ denote the alignment labels for the given sentence pair $\mathbf{x}$ and $\mathbf{y}$. We define the learning objective as:
\begin{equation}
L = \sum_{ij}\hat{A}_{ij}\frac{1}{2} \left(\frac{(S_{\mathbf{x}\mathbf{y}})_{ij}}{n}+ \frac{(S_{\mathbf{y}\mathbf{x}}^{T})_{ij}}{m}\right),
\label{eq:obj_sup}
\end{equation}
where $S_{\mathbf{x}\mathbf{y}}$ and $S_{\mathbf{y}\mathbf{x}}$ are the alignment probability matrices, $n$ and $m$ are the length of sentence $\mathbf{x}$ and $\mathbf{y}$, respectively. Intuitively, this objective encourages the gold aligned words to have closer contextualized representations. In addition, as both $S_{\mathbf{x}\mathbf{y}}$ and  $S_{\mathbf{y}\mathbf{x}}^{T}$ are encouraged to be close to $\hat{A}$, it implicitly encourages the two alignment probability matrices to be symmetrical to each other as well.

Our framework can be easily extended to cases where alignment labels are unavailable, by replacing $\hat{A}$ with pseudo labels $A$ (Equation~\ref{eq:align}) and training in a self-supervised manner. 

\section{Experiments}
\subsection{Setup}
As we aim at building an accurate multilingual word aligner, we evaluate \mname on a diverse alignment test set of seven language pairs: de/sv/ro/fr/ja/zh/fa-en. For finetuning \pnamec, we use nl/cs/hi/tr/es/pt-en as the training set and cs-en as the validation set. To reduce the alignment annotation efforts and the finetuning cost, our training set only contains $3,362$ annotated sentence pairs. To simulate the most difficult use cases where the test language pair may not included in training, we set the test language pairs different from training and validation. Namely, \pname is tested in a zero-shot manner. We denote this dataset as \emph{ALIGN6}.

We induce alignments from 6-th layer of \pnamec, which is selected on the validation set. We use Alignment Error Rate (AER) as the evaluation metric. Our model is not directly comparable to the bilingual baselines, as they build model for each test language pair using large scale parallel corpus of that language pair. In contrast, our method is more efficient as it supports all language pairs in a single model and our finetuning only requires $3,362$ sentence pairs. Appendix~\ref{sec:exp} show more dataset, model, baselines and other setup details.

\subsection{Main Results}
\begin{table}[!t]
\centering
\resizebox{0.8\columnwidth}{!}{
\begin{tabular}{llrr}
\hline
 Model               & & fi-el & fi-he \\ \hline
SimAglin         & noft &69.3   & 85.8   \\\hline
\multirow{3}{*}{AwesomeAlign}   & noft & 69.8 & 84.4  \\
& self-sup ft & 68.8  & 87.7 \\
& sup ft & 67.4 & 86.1 \\\hline
\multirow{3}{*}{\mname} & noft    &  47.0   & 81.2   \\
& self-sup ft   &  40.8   & 76.1    \\ 
& sup ft &  \textbf{\underline{36.7}}   & \textbf{\underline{71.7}}   \\ \hline
\end{tabular}}
\caption{AER comparison between \mname and multilingual baselines on non-English zero-shot language pairs. The best AER for each column is bold and underlined.}
\label{exp:nonen}
\end{table}

Table~\ref{tab:main} shows the comparison of our methods against baselines. \mnamec-supft achieves new SOTA on word alignment induction, outperforming all baselines in $6$ out of $7$ language pairs. \mname is also simpler than AwesomeAlign, which is the best existing multilingual word aligner, as AwesomeAlign finetunes with a combination of five objectives, while \mname only has one objective. The vanilla \pname is a strong multilingual word aligner (see \mnamec-noft). It performs better than SimAlign-noft and AwesomeAlign-noft, and comparable with AwesomeAlign-supft, indicating that \pname has learned high-quality language-agnostic word embeddings. Our finetuning method is effective as well, improving \mnamec-noft by 1.6 and 2.7 AER with self-supervised and supervised alignment labels, respectively. Our model improves multilingual baselines even more significantly on non-English language pairs. See Table~\ref{exp:nonen} of appendix for detailed results.

\subsection{Analysis}\label{sec:ana}
\paragraph{Performance on non-English Language Pair}
We conduct experiments to evaluate \mname against multilingual baselines on non-English test language pairs. The fi-el (Finnish-Greek) and fi-he (Finnish-Hebrew) test set contains 791 and 2,230 annotated sentence pairs, respectively. Both test sets are from \citet{imanigooghari-etal-2021-graph}\footnote{https://github.com/cisnlp/graph-align}. The results are shown in Table~\ref{exp:nonen}. As can be seen, \mname in all three settings significantly improves all multilingual baselines. The improvements is much larger compared with zero-shot English language pairs, demonstrating the effectiveness of \mname on non-English language pairs. We also observe that finetuning better improves \mname than AwesomeAlign. This verifies the strong cross-lingual transfer ability of \pname, even between English-centric and non-English language pairs.

\paragraph{Adapter-based vs. Full Finetuning}
We compare full and adapter-based fine-tuning in Table~\ref{tab:adap}. Compared with full finetuning, adapter-based finetuning updates much less parameters and obtains better performance under both supervised and self-supervised settings, demonstrating its efficiency and effectiveness for zero-shot word alignments.

\begin{table}
\centering
\resizebox{1.0\columnwidth}{!}{
\begin{tabular}{l|l|rr}
\hline
\multicolumn{2}{c|}{Ft type} & full     &  adapter \\ \hline

\multirow{2}{*}{Ft mode} &self-supervised (avg.) & 17.4 & 17.2 \\ \cline{2-4}
&supervised (avg.) & 16.2 & 16.1 \\ \hline
\multicolumn{2}{c|}{Number of ft param.} & 428M & 2.4M\\
\hline

\end{tabular}
}
\caption{AER comparison of full finetuning and adapter-based finetuning.}
\label{tab:adap}
\end{table}

\paragraph{Bilingual Finetuning}\label{para:bift}
\begin{table*}[!t]
\centering
\resizebox{1.6\columnwidth}{!}{\begin{tabular}{ll|rrrrrr}
\hline
Model     & \diagbox{Test lang.}{Ft lang.}  & de-en & fr-en & ro-en & ja-en & zh-en & avg. \\ \hline

\multirow{2}{*}{AwesomeAlign}  & ft lang. & 14.9 & 4.0 & 22.9 & 38.1 & 14.1 & 18.8 \\
& zero-shot langs (avg.)  &16.3 & 4.7 & 26.6 & 43.7 & 15.0 & 21.3 \\
\hline

\multirow{2}{*}{\mname}  & ft lang.  &  14.2 & 3.8 & 21.0 & 38.0 & 13.8 & 18.2 \\ 
& zero-shot langs (avg.) & 14.8 & 3.9 & 20.7 & 40.5 & 13.8 & 18.8 \\ \hline

\end{tabular}}
\caption{AER results with bilingual finetuning.}
\label{tab:bi}
\end{table*}
To better understand our method, we compare with AwesomeAlign under bilingual finetuning setup where the model is finetuned and tested in the same single language pair. We follow the setup in ~\cite{dou-neubig-2021-word} and use finetuning corpus without human-annotated labels. As shown in Table~\ref{tab:bi}, \pname outperforms AwesomeAlign in the finetuning language pair (18.8 vs. 18.2). The performance gap becomes larger for zero-shot language pairs (21.3 vs. 18.8). The results demonstrate that \mname is an effective zero-shot aligner, as \pname has learned more language-agnostic representations which benefit cross-lingual transfer.  

\paragraph{Different Multilingual Pretrained Models}
We investigate the performance of \mnamec-noft when replacing \pname with other mPLMs, including XLM-R, mBERT and four other multilingual sentence Transformer from HuggingFace. LaBSE outperforms other mPLMs by 3.5 to 9.6 averaged AER. Table~\ref{tab:sbert} in appendix shows more details.

\paragraph{Performance across Layer} We investigate the performance of \mnamec-noft when extracts alignments from different layers. Layer $6$, which is the layer we use for all experiments, outperforms other layers by 0.1 to 26.0 averaged AER. Please refer to Table~\ref{tab:layers} in appendix for more details.

\paragraph{Representation Analysis}\label{sec:rep}
To succeed in multilingual word alignment, the contextual embeddings should prefer two following properties: (1) language-agnostic: two aligned bilingual words should be mapped to nearby features in the same language-agnostic feature space. (2) word-identifiable: the embeddings of two random tokens from the same sentence should be distinguishable. 

\begin{figure}[t]
\centering
\includegraphics[width=0.5\textwidth]{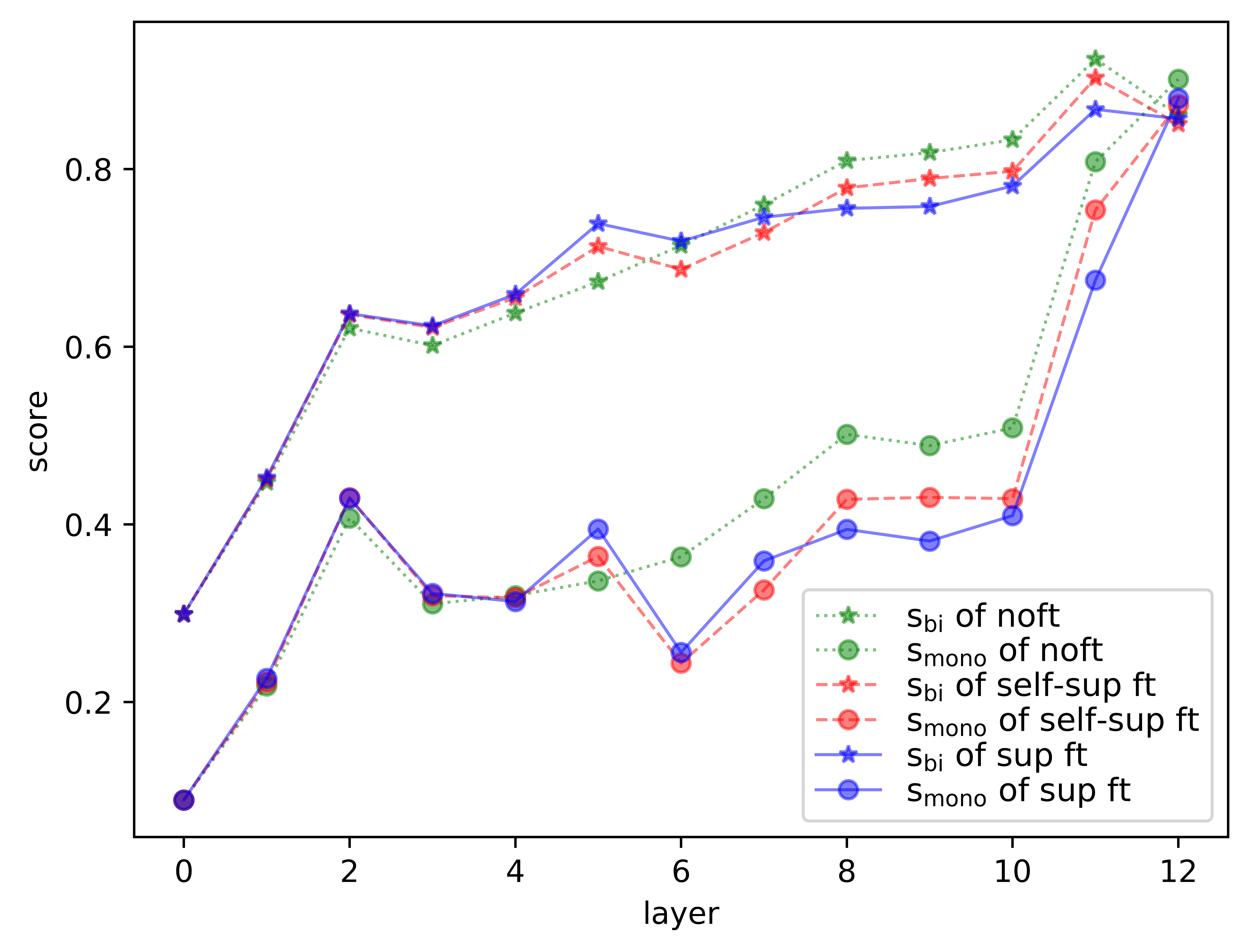}
\caption{$s_{\text{bi }}$ ($\uparrow$) and $s_{\text{mono}}$ ($\downarrow$) of \mname without finetuning (noft), with self-supervised finetuning (self-sup ft) and supervised finetuning (sup ft).} 
\label{fig:analysis}
\end{figure}

Therefore, we analyze the embeddings from different layers of \mname under different settings by computing cosine similarity for aligned word pairs and word pairs randomly sampled from the same sentence, denoted as $s_{\text{bi }}$ and $s_{\text{mono}}$ (see appendix for more experiment details). Intuitively, bigger $s_{\text{bi }}$ and smaller $s_{\text{mono}}$ are preferred as we expect the features of aligned words to be similar while that of two different words to be different. The results on de-en test set are presented in Figure~\ref{fig:analysis}. For vanilla \pname (green curves), we find that features from $6$-th layer, namely the best layer to induce alignment, successfully trades off these two properties as it obtains the biggest $s_{\text{bi }}-s_{\text{mono}}$ among all layers. In addition, adapter-based finetuning improves performance mainly by making features more word-identifiable, as it significantly decreases $s_{\text{mono}}$ while almost maintaining $s_{\text{bi }}$.

\section{Conclusion}
In this paper, we introduce \mnamec, a novel multilingual word aligner based on multilingual sentence Transformer \pnamec. The best proposed approach finetunes \pname on a few thousands of annotated parallel sentences and achieves state-of-the-art performance even for zero-shot language pairs. \mname is believed to be a valuable alignment tool that can be used out-of-the-box for other NLP tasks.

\section*{Limitations}
\mname has shown to extract high quality word alignments when the input texts are two well-paired bilingual sentences. However, the condition is not always met. In lexically constrained decoding of NMT~\cite{hasler2018neural,song2020alignment,chen2021lexically}, the aligner takes a full source-language sentence and a partial target-language translation as the input at each step to determine the right position to incorporate constraints. In creating translated training corpus in zero-resource language for sequence tagging or parsing~\cite{ni-etal-2017-weakly,jain-etal-2019-entity,fei-etal-2020-cross}, the aligner extracts alignments from the labelled sentence and its translation to conduct label projection. Both cases deviate from our current settings as the input sentence may contain translation error or even be incomplete. We leave exploring the robustness of \mname as the future work. 

At the same time, our proposed method only supports languages included in \pnamec. This hinders applying \mname to more low-resource languages. Future explorations are needed to rapidly adapt \mname to new languages~\cite{neubig-hu-2018-rapid,garcia2021towards}.

\section*{Acknowledgements}
This project was supported by National Natural Science Foundation of China (No. 62106138) and Shanghai Sailing Program (No. 21YF1412100). We thank the anonymous reviewers for their insightful feedbacks on this work.

\bibliography{custom}
\bibliographystyle{acl_natbib}

\clearpage
\pagebreak 

\appendix

\section{\pnamec}
LaBSE \cite{feng-etal-2022-language} is the state-of-the-art model for the cross-lingual sentence retrieval task. Given an input sentence, the model can retrieve the most similar sentence from candidates in a different language. It has 471M parameters and supports 109 languages. The model is first pretrained on a combination of masked language modeling \cite{devlin-etal-2019-bert} and translation language modeling \cite{conneau2019cross} tasks on the 17B monolingual data and 6B bilingual translation pairs, respectively. After that, it is effectively finetuned with contrastive loss on 6B bilingual translation pairs across 109 languages. 

Specifically, given a bilingual sentence pair $\langle \mathbf{x}^i,\mathbf{y}^i \rangle$, we use $e_{\mathbf{x}^i}$ and $e_{\mathbf{y}^i}$ to denote their sentence embeddings from LaBSE. Then the model is finetuned using contrative loss with in-batch negatives~\cite{chen2020simple}:
\begin{align*}
\ell=&-\frac{1}{N} \sum_{i=1}^{N} \left\{\log \frac{\exp \left(\phi(e_{\mathbf{x}^i}, e_{\mathbf{y}^i})\right)}{\sum_{j=1}^{N} \exp \left(\phi(e_{\mathbf{x}^i}, e_{\mathbf{y}^j})\right)}+ \right.
\\
&\left. \log \frac{\exp \left(\phi(e_{\mathbf{x}^i}, e_{\mathbf{y}^i})\right)}{\sum_{j=1}^{N} \exp \left(\phi(e_{\mathbf{x}^j}, e_{\mathbf{y}^i})\right) } \right\}, \numberthis
\label{eq:cons}
\end{align*}
where $\phi(e_{\mathbf{x}^i}, e_{\mathbf{y}^j})$ measures the similarity of sentence $\mathbf{x}^i$ and $\mathbf{y}^j$ in the embedding space: 
\begin{align*}
\phi\left(e_{\mathbf{x}^i}, e_{\mathbf{y}^j}\right)= \begin{cases}e_{\mathbf{x}^i}^\top e_{\mathbf{y}^j}-b & \text { if } i=j \\ e_{\mathbf{x}^i}^\top e_{\mathbf{y}^j} & \text { if } i \neq j\end{cases}. \numberthis
\end{align*}
Note that a margin $b$ is introduced to improve the separation between positive and negative pairs. 

\section{Experiments Setup}\label{sec:exp}

\subsection{Language Code}
We refer to the language information in Table 1 of ~\cite{fan2021beyond}. The information of the languages used in this paper is listed in Table~\ref{tab:lang}.

\begin{table}
\centering
\begin{tabular}{llll}
\hline
ISO & Name &Family  \\ \hline
en   & English    & Germanic  \\
nl    & Dutch     & Germanic  \\
cs  & Czech     & Slavic \\
hi    & Hindi     & Indo-Aryan  \\
tr    & Turkish     & Turkic \\
es    &  Spanish   & Romance   \\
pt    & Portuguese & Romance     \\
de    & German   & Germanic   \\
sv    & Swedish     & Germanic  \\
fr    & French   & Romance   \\
ro    & Romanian    & Romance   \\
ja    & Japanese   & Japonic   \\
zh    & Chinese     & Chinese  \\
fa    & Persian & Iranian    \\
 \hline

\end{tabular}
\caption{The information of the languages used in this
paper. }
\label{tab:lang}
\end{table}

\begin{table*}[!t]
\centering
\resizebox{1.0\textwidth}{!}{
\begin{tabular}{lllp{9.0cm}r}
\hline
Type & Lang. & Source & Link &$\#\;$Sents  \\ \hline 
\multirow{7}{*}{Training set}
&cs-en       & \citet{marevcek2011automatic}    & \url{http://ufal.mff.cuni.cz/czech-english-manual-word-alignment} & 2400    \\
&nl-en       & \citet{macken2010annotation}     & \url{http://www.tst.inl.nl} & 372    \\
&hi-en           & \citet{aswani2005aligning}    & \url{http://web.eecs.umich.edu/~mihalcea/wpt05/} & 90    \\
&tr-en          & \citet{cakmak2012word}    & \url{http://web.itu.edu.tr/gulsenc/resources.htm} & 300    \\
% &es-en         & \citet{graca2008building}     & \url{http://www.l2f.inesc-id.pt/resources/translation/}&  100    \\
&es-en         & \citet{graca2008building}     & \url{https://www.hlt.inesc-id.pt/w/Word_Alignments} &  100  \\
&pt-en          & \citet{graca2008building}   & \url{https://www.hlt.inesc-id.pt/w/Word_Alignments}  & 100    \\ \hline
Validation set & cs-en           & \citet{marevcek2011automatic}   & \url{http://ufal.mff.cuni.cz/czech-english-manual-word-alignment}  & 101    \\ \hline
\multirow{7}{*}{Test set} 
& de-en        & \citet{vilar2006aer}   & \url{http://www-i6.informatik.rwth-aachen.de/goldAlignment/}  & 508    \\
&sv-en         & \citet{holmqvist2011gold}     & \url{https://www.ida.liu.se/divisions/hcs/nlplab/resources/ges/} &192    \\
&fr-en          & \citet{mihalcea-pedersen-2003-evaluation}   & \url{http://web.eecs.umich.edu/~mihalcea/wpt/}  & 447    \\
&ro-en           & \citet{mihalcea-pedersen-2003-evaluation} & \url{http://web.eecs.umich.edu/~mihalcea/wpt05/}    & 248    \\
&ja-en          & \citet{neubig11kftt}     & \url{http://www.phontron.com/kftt} &582    \\
&zh-en           & \citet{liu2015contrastive}     & \url{https://nlp.csai.tsinghua.edu.cn/~ly/systems/TsinghuaAligner/TsinghuaAligner.html} &450    \\
&fa-en       & \citet{tavakoli2014phrase}  & \url{http://eceold.ut.ac.ir/en/node/940}   & 400    \\\hline
\end{tabular}
}
\caption{Training, validation and test dataset of ALIGN6. Note that this is a zero-shot setting as the test language pairs do not appear in training and validation.}
\label{tab:data}
\end{table*}

\subsection{Dataset}
\label{sec:appendix}
Table~\ref{tab:data} shows the detailed data statistics of ALIGN6. The ja and zh sentences are preprocessed by~\citet{dou-neubig-2021-word} and \citet{liu2015contrastive}, respectively. For finetuning \mname and multilingual baselines, we use the training and validation set from ALIGN6. As bilingual baselines are not capable of zero-shot alignment induction, they are trained from scratch with parallel corpus of the test language pair using the same dataset as~\citet{dou-neubig-2021-word}. The bilingual training data set of de/fr/ro/ja/zh-en contain 1.9M, 1.1M, 450K, 444K and 40K parallel sentence pairs, respectively, which are much larger than the training dataset of ALIGN6.

\subsection{Model Setup}
We use the contextual word embeddings from the 6-th layer of the official \pnamec\footnote{https://huggingface.co/sentence-transformers/LaBSE}, which have 768 dimensions. We set the threshold in Equation~\ref{eq:align} to 0.1, which is selected on validation set by manual tuning among $[0,0.2]$. For adapter-based finetuning, we set the hidden dimension of the adapters to be 128. The adapters have 2.4M parameters, which account 0.5\% of the parameters of \pnamec. We use the AdamW optimizer with learning rate of 1e-4, and do not use warmup or dropout. The batch size is set to 40 and maximum updates number is 1500 steps. We use a single NVIDIA V100 GPU for all experiments.

\subsection{Baselines}
Besides three statistical baselines fast-align~\cite{dyer-etal-2013-simple}, eflomal~\cite{ostling2016efficient} and GIZA++~\cite{och-ney-2003-systematic}, we compare \mname with the following neural baselines:

\noindent \textbf{MTL-FULLC-GZ}~\cite{garg-etal-2019-jointly}. This model supervises an attention head in Transformer-based NMT model with GIZA++ word alignments in a multitask learning framework.

\noindent \textbf{BAO-GUIDE}~\cite{zenkel-etal-2020-end}. This model adds an extra alignment layer to repredict the
to-be-aligned target token and further improves performance with Bidirectional Attention Optimization.

\noindent \textbf{SHIFT-AET}~\cite{chen-etal-2020-accurate}. This model trains a separate alignment module in a self-supervised manner, and induce alignments when the to-be-aligned target token is the decoder input.

\noindent \textbf{MASK-ALIGN}~\cite{chen-etal-2021-mask}. This model is a self-supervised word aligner which makes use of the full context on the target side. 

\noindent \textbf{BTBA-FCBO-SST}~\cite{zhang-van-genabith-2021-bidirectional}. This model has similar idea with ~\citet{chen-etal-2021-mask}, but with different model architecture and training objectives.

\noindent \textbf{SimAlign}~\cite{jalili-sabet-etal-2020-simalign}. This model is a multilingual word aligner which induces alignment with contextual word embeddings from mBERT and XLM-R. 

\noindent \textbf{AwesomeAlign}~\cite{dou-neubig-2021-word}. This model improves over SimAlign by designing new alignment induction method and proposing to further finetune the mPLM on parallel corpus. 

Among them, SimAlign and AwesomeAlign are multilingual aligners which support multiple language pairs in a single model, while others are bilingual word aligners which require training from scratch with bilingual corpus for each test language pair. We re-implement SimAlign and AwesomeAlign, while quote the results from \cite{dou-neubig-2021-word} for the three statistical baselines and the corresponding paper for other baselines.

\subsection{Sentence Transformer}
\label{sec:sbert}
We compare \pname with four other multilingual sentence Transformer in HuggingFace. The detailed information of these models are:

\noindent \textbf{distiluse-base-multilingual-cased-v2}.\footnote{https://huggingface.co/sentence-transformers/distiluse-base-multilingual-cased-v2} This model is a multilingual knowledge distilled version of m-USE~\cite{yang2020multilingual}, which has 135M parameters and supports more than 50+ languages.

\noindent \textbf{paraphrase-xlm-r-multilingual-v1}.\footnote{https://huggingface.co/sentence-transformers/paraphrase-xlm-r-multilingual-v1} This model is a multilingual version of paraphrase-distilroberta-base-v1~\cite{reimers-2019-sentence-bert}, which has 278M parameters and supports 50+ languages. It initializes the student model with an mPLM and trains it to imitate monolingual sentence Transformer on parallel data with knowledge distillation.

\noindent \textbf{paraphrase-multilingual-MiniLM-L12-v2}.\footnote{https://huggingface.co/sentence-transformers/paraphrase-multilingual-MiniLM-L12-v2} This model is a multilingual version of paraphrase-MiniLM-L12-v2~\cite{reimers-2019-sentence-bert}, which has 118M parameters and supports 50+ languages. It trains similarly as paraphrase-xlm-r-multilingual-v1, but with different teacher and student model initialization.

\noindent \textbf{paraphrase-multilingual-mpnet-base-v2}.\footnote{https://huggingface.co/sentence-transformers/paraphrase-multilingual-mpnet-base-v2} This model is a multilingual version of paraphrase-mpnet-base-v2~\cite{reimers-2019-sentence-bert}, which has 278M parameters and supports 50+ languages. It trains similarly as paraphrase-xlm-r-multilingual-v1, but with different teacher model initialization.

\subsection{Bilingual Finetuning}
We use the same dataset as bilingual baselines for bilingual finetuning following~\cite{dou-neubig-2021-word}. At each time, we finetune \pname with one language pair among de/fr/ro/ja/zh-en and test on all seven language pairs. For Awesome-align, we follow the setup in their paper, while for \mnamec, we use the same hyperparameters as the main experiments.

\subsection{Representation Analysis}
We conduct representation analysis on de-en test set. To compute $s_{\text{bi}}$, we calculate the averaged cosine similarity of all gold aligned bilingual word pairs. To compute $s_{\text{mono}}$, we randomly permute a given sentence $\mathbf{x}=\langle x_1, x_2, ...,x_n \rangle$ to get $\mathbf{x}^{\prime}=\langle x_1^{\prime}, x_2^{\prime}, ...,x_n^{\prime} \rangle$ and then create n word pairs as $\{\langle x_i\text{-}x_i^{\prime}\rangle\}_{i=1}^n$. We go through all de and en test sentences and report the averaged cosine similarity of all created word pairs as $s_{\text{mono}}$.

\section{Experiment Results}
Detailed results for each test language in Section~\ref{sec:ana} are shown in Table~\ref{tab:adapter} to Table~\ref{tab:layers}.

\begin{table*}
\centering
\resizebox{2.0\columnwidth}{!}{
\begin{tabular}{ll|lllllllllr}
\hline
Ft mode &Ft type                             &  & de-en & sv-en & fr-en & ro-en & ja-en & zh-en  & fa-en &avg \\ \hline

\multirow{2}{*}{Self-supervised } &full                        &  & 14.7 & 5.8 & 3.7  & 21.6 & 39.9 & 13.3  & 22.7 & 17.4 \\ 

& adapter                      &  & 14.3 & 5.8 & 3.9 & 21.6 & 39.2 & 13.0  & 22.6 & 17.2 \\ \hline

\multirow{2}{*}{Supervised } & full                       &  & \textbf{\underline{13.6}} & 5.3 & 2.8  & 21.0 & 37.1 & \textbf{\underline{11.0}}  & 22.5 & 16.2 \\ 

& adapter                      &  & \textbf{\underline{13.6}} & \textbf{\underline{5.2}} & \textbf{\underline{2.7}} & \textbf{\underline{20.8}} & \textbf{\underline{36.8}} & 11.5  & \textbf{\underline{22.2}} & \textbf{\underline{16.1}} \\

\hline

\end{tabular}
}
\caption{AER comparison of full finetuning and adapter-based finetuning. The best AER for each column is bold and underlined. }
\label{tab:adapter}
\end{table*}

\begin{table*}[!t]
\centering
\resizebox{1.7\columnwidth}{!}{\begin{tabular}{ll|rrrrrrr}
\hline
Model     & \diagbox{Ft lang.}{Test lang.}  & de-en & fr-en & ro-en & ja-en & zh-en & sv-en & fa-en \\ \hline

\multirow{5}{*}{AwesomeAlign}  & de-en & \textbf{\underline{14.9}} & 4.7 & 26.2 & 43.6 & 14.6 & 7.1 & 28.2 \\ 
& fr-en & 16.4 & \textbf{\underline{4.0}} & 26.9 & 44.6 & 15.7 & 7.6 & 28.0 \\
& ro-en & 15.8 & 4.7 & \textbf{\underline{22.9}} & 44.2 & 15.1 & 7.8 & 27.0 \\
& ja-en & 16.8 & 4.9 & 27.0 & \textbf{\underline{38.1}} & 15.2 & 8.5 & 30.0 \\
& zh-en & 16.2 & 4.6 & 26.2 & 42.4 & \textbf{\underline{14.1}} & 8.1 & 28.0 \\ \hline

\multirow{5}{*}{\mname}  & de-en & \textbf{\underline{14.2}}  & 3.8 & 20.9 & 39.3 & 13.1 & 5.7 & 22.5  \\ 
& fr-en & 14.6 & \textbf{\underline{3.8}} & 20.8 & 41.0 & 14.1 & 6.0 & 22.5 \\
& ro-en & 15.2 & 4.0 & \textbf{\underline{21.0}} & 42.1 & 14.4 & 6.5 & 23.2 \\
& ja-en & 14.8 & 3.9 & 20.3 & \textbf{\underline{38.0}} & 13.5 & 6.3 & 22.5 \\
& zh-en & 14.6 & 3.9 & 20.7 & 38.9 & \textbf{\underline{13.4}} & 5.9 & 22.4 \\ \hline

\end{tabular}}
\caption{AER results with bilingual finetuning. The results where the model is trained and tested on the same language pair are bold and underlined.}
\label{tab:bilingual}
\end{table*}

\begin{table*}[!t]
\centering
\resizebox{2.0\columnwidth}{!}{
\begin{tabular}{l|r|rrrrrrrrr}
\hline
      & layer   &  de-en & sv-en & fr-en & ro-en & ja-en & zh-en  & fas-en & avg \\ \hline 
mBERT & 8        & 17.4 & 8.7 & 5.6      & 27.9       & 45.6  & 18.1  & 33.0 & 22.3  \\ 
XLM-R & 8       & 23.1 & 13.3 & 9.2      & 28.6      & 62.0  & 30.3   & 28.6 & 27.9 \\\hline

distiluse-base-multilingual-cased-v2    
      & 3       & 23.7   & 17.2   & 9.8  & 29.2    & 56.3 &29.2      & 33.5   & 28.4   \\ 

paraphrase-xlm-r-multilingual-v1     
      & 6       & 17.4 & 8.7 & 4.9       & 24.7       & 53.8 & 26.1   & 26.5 & 23.2      \\ 

paraphrase-multilingual-MiniLM-L12-v2     
      & 6       & 19.4 & 9.4 & 6.2       & 26.0       & 57.7 & 29.7   & 27.4 & 25.1      \\ 
paraphrase-multilingual-mpnet-base-v2     
      & 5       & 18.0 & 8.9 & 5.4       & 24.1      & 54.9 & 25.7   & 25.5  & 23.2     \\ 
LaBSE & 6       & \textbf{\underline{16.0}} & \textbf{\underline{ 7.3}} & \textbf{\underline{ 4.5}}      & \textbf{\underline{ 20.8}}   & \textbf{\underline{ 43.3}} & \textbf{\underline{ 16.2}} & \textbf{\underline{ 23.4}} & \textbf{\underline{ 18.8}}     \\\hline
      
\end{tabular}}
\caption{AER comparison of \pname and other multilingual pretrained model. All are without finetuning. We determine the best layer of alignment induction for each model using the validation set. The best AER for each column is bold and underlined.}
\label{tab:sbert}
\end{table*}

\begin{table*}[!t]
\centering
\resizebox{1.5\columnwidth}{!}{
\begin{tabular}{l|rrrrrrrrrr}
\hline
     Layer &  &  de-en & sv-en & fr-en & ro-en & ja-en & zh-en  & fa-en & avg\\ \hline 
     0      &     & 32.4 & 27.7 & 20.5      & 44.2      & 65.5  & 40.1  & 38.7 & 38.4   \\ 
     1   &        & 27.3 & 19.7 & 12.8      & 35.6      & 64.0 & 33.9  & 35.4 & 32.7    \\
     2   &        & 22.3 & 14.0 & 8.6      & 28.8      & 58.0 & 25.0  & 31.3  & 26.9  \\ 
     3   &        & 18.5 & 9.9 & 6.0      & 24.0      & 50.3 & 17.9  & 26.8  & 21.9  \\ 
     4   &        & 17.7 & 8.7 & 5.9      & 23.3      & 48.4 & 16.3  & 25.7  & 20.9  \\ 
     5   &        & \textbf{\underline{15.8}} & 7.4 & \textbf{\underline{4.5}}      & 21.5      & 43.7 & 15.4  & 23.8  & 18.9  \\
     6   &        & 16.0 & \textbf{\underline{7.3}} & \textbf{\underline{4.5}}      & \textbf{\underline{20.8}}      & 43.3 & 16.2   & 23.4  & \textbf{\underline{18.8}}  \\
     7   &        & 16.5 & 7.6 & 4.8      & 22.4      & 43.4 & \textbf{\underline{15.0}}  & 23.7  & 19.1  \\
     8   &        & 16.2 & \textbf{\underline{7.3}} & 5.0      & 21.6      & \textbf{\underline{42.7}} & 16.7  & 23.4  & 19.0  \\ 
     9   &        & 16.8 & 7.6 & 5.3      & 21.5      & \textbf{\underline{42.7}} & 17.9  & \textbf{\underline{23.2}}  & 19.3  \\ 
     10   &        & 17.7 & 9.0 & 5.6      & 23.0      & 44.4 & 20.4  & 24.4 & 20.6   \\ 
     11   &        & 36.7 & 27.0 & 24.2      & 43.6      & 61.3 & 35.0  & 46.2 & 39.1   \\ 
     12   &        & 43.1 & 33.2 & 30.5      & 46.0      & 65.7 & 42.6  & 52.4 & 44.8   \\ \hline
\end{tabular}}
\caption{AER comparison of vanilla \pname across layers. Layer 0 is the embedding layer. The best AER for each column is bold and underlined.}
\label{tab:layers}
\end{table*}

\end{document}